\pdfoutput=1

\documentclass[10pt,a4paper,twocolumn]{article}
\usepackage[utf8]{inputenc}
\usepackage{mathtools} %\usepackage{amsmath}
\usepackage{amsfonts}
\usepackage{amssymb}
\usepackage{bm}
\usepackage{import} 
\usepackage{graphicx}
\usepackage{xcolor}
\usepackage[nolist,nohyperlinks]{acronym}
\usepackage{tabulary}
\usepackage{booktabs}
\usepackage{siunitx}
\usepackage[backend=biber,sorting=none]{biblatex}
\usepackage{soul} %for highlighting

\providecommand{\keywords}[1]{\textbf{\textit{Index terms---}} #1}

\author{Miguel Sim\~ao\thanks{M. Sim\~ao is with the Department of Mechanical Engineering of the University of Coimbra, Portugal (miguel.simao@uc.pt).}, Pedro Neto\thanks{P. Neto is with the Department of Mechanical Engineering of the University of Coimbra, Portugal (pedro.neto@dem.uc.pt).} and Olivier Gibaru\thanks{O. Gibaru is with the École Nationale Supérieure d'Arts et Métiers, Lille, France (olivier.gibaru@ensam.eu).}%
}

\title{Improving Novelty Detection with Generative Adversarial Networks on Hand Gesture Data}

\begin{document}
	
	\maketitle
	\begin{abstract}
		We propose a novel way of solving the issue of classification of out-of-vocabulary gestures using \acp{ANN} trained in the \ac{GAN} framework. A generative model augments the data set in an online fashion with new samples and stochastic target vectors, while a discriminative model determines the class of the samples. The approach was evaluated on the UC2017 SG and UC2018 DualMyo data sets. The generative models' performance was measured with a distance metric between generated and real samples. The discriminative models were evaluated by their accuracy on trained and novel classes. In terms of sample generation quality, the \ac{GAN} is significantly better than a random distribution (noise) in mean distance, for all classes. 
		In the classification tests, the baseline neural network was not capable of identifying untrained gestures. When the proposed methodology was implemented, we found that there is a trade-off between the detection of trained and untrained gestures, with some trained samples being mistaken as novelty. Nevertheless, a novelty detection accuracy of 95.4\% or 90.2\% (depending on the data set) was achieved with just 5\% loss of accuracy on trained classes.
	\end{abstract}

	\keywords{Collaborative Robotics, Semi-Supervised Learning, Generative Adversarial Networks, Novelty Detection}
	
	% !TeX spellcheck = en_GB
% !TeX encoding = UTF-8

%List of acronyms used in this paper:
\begin{acronym}[AC-GAN]
	%\acro{}{}	
	\acro{AC-GAN}{Auxiliary Conditional Generative Adversarial Network}
	\acro{ANN} {Artificial Neural Network}
	\acro{BP}  {Backpropagation}
	\acro{CNN} {Convolutional Neural Network}
	\acro{DBN} {Deep Belief Network}
	\acro{DDR} {Data Dimensionality Reduction}
	\acro{DG}  {Dynamic Gesture}
	\acro{DOF} {Degree of Freedom}
	\acrodefplural{DOF}{Degrees of Freedom}
	\acro{DTW} {Dynamic Time Warping}
	\acro{EMG} {Electromyography}
	\acro{FFNN}{Feed-Forward Neural Network}
	\acro{GAN} {Generative Adversarial Network}
	\acro{HCI} {Human-Computer Interaction}
	\acro{HMI} {Human-Machine Interaction}
	\acro{HMM} {Hidden Markov Model}
	\acro{HRC} {Human-Robot Collaboration}
	\acro{HRI} {Human-Robot Interaction}
	\acro{IMU} {Inertial Measurement Unit}
	\acro{LSTM}{Long Short-Term Memory}
	\acro{ME}  {Movement Epenthesis}
	\acro{MLNN}{Multi-Layer Neural Network}
	\acro{MLP} {Multi-Layer Perceptron}
	\acro{ND}  {Novelty Detection}
	\acro{NLP} {Natural Language Processing}
	\acro{PC}  {Principal Component}
	\acro{PCA} {Principal Component Analysis}
	\acro{RNN} {Recurrent Neural Network}
	\acro{RR}  {Recognition Rate}
	\acro{SCG} {Scaled Conjugate Gradient}
	\acro{sEMG}{Surface Electromyography}
	\acro{SG}  {Static Gesture}
	\acro{SGD} {Stochastic Gradient Descent}
	\acro{SME} {Small and Medium-sized Enterprises}
	\acro{SOI} {Segment of Interest}
	\acrodefplural{SOI}{Segments of Interest}
	\acro{SVD} {Singular Value Decomposition}
	\acro{SVM} {Support Vector Machine}
	
\end{acronym}	
	% !TeX encoding = UTF-8
% !TeX spellcheck = en_GB
% !TeX root = paper-gan-master.tex

% chap4-gan.tex
%
% GENERATIVE ADVERSARIAL NETWORKS FOR TIME-SERIES CLASSIFICATION:
%	- Introduction
%		- What are the problems?
%		- Possible solutions
%		- GAN
%	- Methods and Methodology
%	- Results

%-------------------------------------------------------------------------------

%\chapter{Generative Adversarial Networks for Gesture Data}

\section{Introduction}

Often-times the performance of a classifier trained offline on a data set is not indicative of the online performance. This may happen due to missing elements on the data processing pipeline, such as proper data scaling. However, the major issue is the limited scope of a data set, when compared to the real problem. Independently of the resources available, a data set can not include a large portion of real-world scenarios. In those cases, there is no way of confidently predicting the performance of a classifier.

In a gesture recognition data set, there are training patterns of a predefined number of gesture classes. It is also possible to include gesture patterns that do not match any of the classes, {\color{black} which is representative of real-world conditions \cite{Neto2013a}}. These are known as untrained gestures, novelties, non-gestures, or \emph{others}, i.e., gestures that do not belong to the predefined classes. However, the diversity of non-gestures is almost infinite, or at least, much greater than that of the predefined gestures.

Non-gestures appear in real-world conditions on every type of interaction, particularly in human-human and human-machine interaction. They may occur due to, albeit not limited to the following reasons:

\begin{enumerate}
	\item The user is uneducated in respect to the human-machine interface and performs out of vocabulary gestures;
	\item The user is distracted and is moving in a way that does not make sense in the context of the interface, e.g., talking to somebody else;
	\item The user is forced to move in response to other elements in the surroundings, e.g., moving machines or falling objects;
	\item The user {\color{black} is moving from} the end of an interaction to the start of the next, i.e., \ac{ME}.
\end{enumerate}

The naive way to exclude non-gestures is to set a threshold to the output probability of a classifier:
\begin{equation}
	\text{class}=
	\begin{dcases}
		\tau, & p(y_\tau|z)\geq\text{threshold} \\
		\text{none}, & p(y_\tau|z) < \text{threshold} \\
	\end{dcases}
	\label{eq:gan_prob_threshold}
\end{equation}
where $y$ is the classifier's output given the feature vector $z$ and $p$ the probability distribution over the problem's classes. We have previously shown that for the most widely used classifier, \acp{ANN}, the output probability is not a good measure of classification certainty \cite{Simao2016}. This means that many correct classifications may have low probabilities and incorrect ones have high likelihood, as predicted by the \ac{ANN}. If we were to set a threshold on the class probability using (\ref{eq:gan_prob_threshold}), many good classifications would be discarded, while an equal proportion of bad classifications would pass. Therefore, the threshold method is not effective for its purpose.

It is possible to exclude non-gestures and {\color{black} miss-classifications} using context clues, such as limiting the quantity of possible outputs. However, this is a limited approach and we are interested in exploring new methodologies that may help a classifier discriminate non-gestures. {\color{black} This problem was formally described by the Neyman and Pearson method (N-P classification), which minimizes the error for a class while keeping the error constant for other classes \cite{Neyman1933}. N-P classification can be implemented by knowing the probability density function of the input samples. More recently, methods based on radial basis function (RBF) neural networks demonstrated better performance than traditional threshold methods (samples near decision boundaries are treated differently from other samples) \cite{Casasent2003}.}

A non-gesture is an "abnormal" occurrence for the classification model, which is trained with a restricted number of classes. While non-gestures could belong to an extra class containing all of the possible non-gestures, training it is a challenge because of the lack of training examples. {\color{black} Owing to the} large gesture domain, it is unfeasible to get data samples from every possible non-gesture. This problem is seldom addressed and most data sets do not include such patterns.

The problem of detecting "abnormal" patterns  from a predefined number of gesture classes is generally known in the literature as \ac{ND}. In this type of problem, the predefined classes have considerably more training examples, while the "abnormal" patterns are under-represented. In \cite{Pimentel2014}, the authors concluded that currently, there is no optimal solution for the \ac{ND} problem because it depends on the type of data and the application domain. Distance methods such as k-NN have been shown to be superior, but their computational efficiency decreases with data set size, therefore making them unsuited for larger data sets and real-time applications \cite{Ding2014}.

We propose the use of a semi-supervised methodology which uses the labelled samples of a data set and generated unlabelled samples which correspond to either gestures or non-gestures. This is an approach that is currently used in deep learning where very large data sets are required but labels are not always available \cite{Kingma2014,Springenberg2015}. This is a methodology that has been used successfully on data sets for image recognition. In this chapter, we present the results of its application to hand-gesture recognition with \acp{GAN}.

\subsection{Generative Adversarial Networks}

The name \ac{GAN} describes a framework for the training of generative neural networks that was introduced by Goodfellow et \emph{al.} in \cite{Goodfellow2014}. In this framework, there are two competing nets which are trained simultaneously, a generative net $G$ and a discriminative net $D$. The objective of the discriminator $D$ is calculating the probability that a sample came from the real data set rather than from the generator $G$. On the other hand, $G$ is trained to produce samples that maximize the probability of $D$ classifying them as real. $D$ and $G$ have competing objectives and should normally improve one another. A diagram representing a variant of this framework is shown in {\color{black} figure \ref{fig:gan_custom_acgan_training}}.

Current applications of \acp{GAN} are essentially in the field of image to image translation, i.e., generation of new images with specified features or increased resolution \cite{Han2017,Huang2018,Li2017,Mao2018,Yuan2018b}. Very few authors have studied applications on other domains, such as speech \cite{Saito2018,Yuan2018} and text generation \cite{Li2018}. The flexibility of \acp{GAN} gave rise to a plethora of network structures and training methods, of which some notable ones are: \ac{AC-GAN} \cite{Odena2016}, Cycle-Consistent \ac{GAN} (CycleGAN) \cite{Zhu2017} and Wasserstein \ac{GAN} \cite{Arjovsky2017}. In terms of performance, it is difficult to evaluate these models quantitatively, since these are generative models and they are purpose built. Nevertheless, the generated outputs are often indistinguishable from real data in state of the art implementations.

The original \acp{GAN} \cite{Goodfellow2014} had a discriminator $D$ whose output was the binary classification of the source of a sample (real or generated). If the original data set was also divided in N classes, the $D$ would also be able to classify generated data, thus appearing an extension to semi-supervised learning \cite{Odena2016}. These are the auxiliary conditional \acp{GAN}.

For the evaluation of the generated samples, an intra-class diversity measure was proposed in \cite{Odena2016}, specifically the multi-scale structural similarity (MS-SSIM). This metric aims to account only for the same features as a human would perceive, rather than calculate a pixel to pixel distance. Thus, it is more indicated to measure image similarity. Lack of similarity is a symptom of an important failure mode during \ac{GAN} training. It happens when the generator collapses and generates a single pattern that maximally confuses the discriminator. A generator model that only outputs a single pattern is very limited and not useful, so we should avoid a collapsed generator.

In this chapter, we propose some modifications to the \ac{AC-GAN} structure: 
\begin{enumerate}
	\item A softmax layer as the second output of the discriminator;
	\item A one-hot encoded second input in the generator instead of the class number;
	\item Training with stochastic target vectors.
\end{enumerate}
These modifications aim to allow the use of the discriminator as an online classifier and the generation of samples with any given class likelihood.

\section{Methods and Methodology}

In this section we discuss the data pipeline for the fitting and test of the model, the architecture of the model, its training methodology and the definition of the tests.

\subsection{Data Pipeline}
\label{sec:gan_sub_data}

We assume that we have available a labelled data set, which is defined as:
\begin{equation}
	\mathcal{D}=\left\{ \left(\bm{X}^{(i)},\iota^{(i)}\right)\ :\quad i=1,2,\ldots,n_{samples} \right\}
\end{equation}
in which $\bm{X} \subset \mathbb{M}^{t \times d}$ represents the sample data of $d$ channels (variables) and $t$ time steps, and $\iota^{(i)} \in \mathbb{N}_0$ is the target class for that sample. For development and testing purposes, the data set is split into three subsets: training, validation and testing subsets. 

The following step is feature extraction, which depends on the data set, classifier model and objectives. Generally speaking, it is defined by $\bm{f}^{(i)} = \mathcal{F}\left ( \bm{X}^{(i)}\right )$, where $\mathcal{F}$ represents the extraction function and $\bm{f}\subset \mathbb{M}^{n_f}$ are the output features, which is a vector of length equal to the number of features per sample, $n_f$.

The features are normalized, i.e., the variables are assumed to follow normal distributions, whose parameters are calculated on the training set. All of the subsets are normalized with these parameters and every new sample is normalized with these same parameters. Finally, the targets are one-hot encoded.

\subsection{Stochastic Target GANs}

%\begin{figure}
%	\centering
%	%\def\svgwidth{\columnwidth}
%	%\input{image_custom_acgan_structure.pdf_tex}
%	%\caption{Diagram representing the custom \ac{AC-GAN} framework. The generator G has two inputs: noise with a latent size $l$ and class, a one-hot encoded vector for $n_c$ classes. Its output is the generated sample, a vector with $n_f$ variables -- the same as the number of features of a real sample. The discriminator D takes as inputs a sample, real or generated, and determines a validity scalar for binary classification of the sample's source. Its second output is a vector with the classification of the sample. $b$ corresponds to the batch size for the gradient descent optimization.}
%	\label{fig:gan_custom_acgan}
%\end{figure}

The structure of the custom \ac{GAN} is very similar to a \ac{AC-GAN} \cite{Odena2016}. There are still separated generator and discriminator networks, such as in the original \ac{GAN} introduced in \cite{Goodfellow2014}. The first input of the generator $G$ is a noise vector $z$ with latent size $l$ that is sampled from a normal distribution $\mathcal{N}\sim \left(\mu=0,\sigma^2=1 \right)$. The second input of $G$ is a one-hot vector that represents the class to be generated, while a \ac{AC-GAN} uses the class index. The generator network's structure is free, but it must be a feed-forward neural network. The structure is highly dependent on the available data and the training hyperparameters.

The input of the discriminator $D$ are vectors of length $n_f$ (number of features), which can either come from data set samples or the generator. The discriminator network's structure is also free and depends on the type of training data. $D$ has two outputs, being the first the validity of the input sample. The validity is a scalar $v \in [0,1] \subset \mathbb{R}$, i.e., a real number between 0 and 1. Validity below $0.5$ denotes a generated sample and if it is above or equal to 0.5, it corresponds to a real sample. The discriminator has a second output, which is the classification of the sample as a one-hot vector.

The real data set samples are represented by a 3-element tuple $x_r,t_r,s_r$, which are the sample data, the target class and its source, respectively. The generated samples are presented by $x_g,t_g,s_g$. The variables $x_r$ and $x_g$ {\color{black} have the same shape as} $\bm{f}$ (section \ref{sec:gan_sub_data}) and represent the features obtained from the corresponding sample. The targets $t_r$ and $t_g$, for real and generated samples respectively, are one-hot encoded vectors of the classes they represent. Assuming the classification problem has $n_c$ classes, the one-hot vector $t$ for class $\iota$ is a horizontal vector of size $1\times n_c$ composed of zeros, except $t_\iota=1$. The source scalar, $s$, is either 1 for real samples, or 0 for generated samples.

\subsection{Model Training} \label{subsec:gan_training}

\begin{figure*}
	\centering
	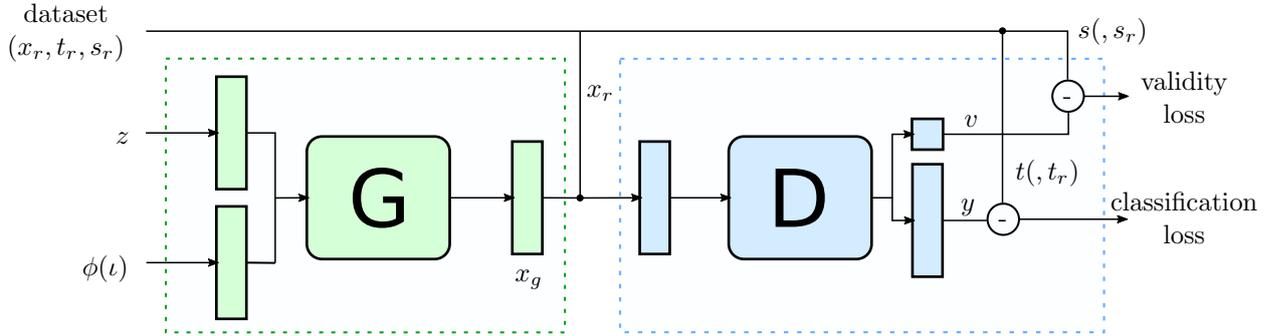
	\caption{Diagram representing the training process of the custom \ac{GAN}.}
	\label{fig:gan_custom_acgan_training}
\end{figure*}

The two networks are trained by \ac{SGD} simultaneously, in two interwoven stages ($D$-$G$-$D$-$G$-$D$-$G\ldots$) for a number of epochs. A diagram that applies to both stages is shown in figure \ref{fig:gan_custom_acgan_training}. In \ac{SGD}, the weights of the model are updated according to the gradient of the model's loss on a batch of $b$ samples. This means that instead of vectors, the inputs and outputs of the networks are matrices, where the first dimension corresponds to the sample index and the second to the variables (feature and target indices).

\subsubsection*{First stage: discriminator}

The discriminator is trained with both real and generated samples. Given a pre-selected batch size $b$, $b/2$ samples are extracted from the data set, denoted as $x_r$. An equal amount of samples is generated from $G$, $x_g$. These samples are generated by running $G$ with two inputs:
\begin{equation}
x_g = G(z,\phi(\iota))
\label{eq:gan_d}
\end{equation}
The first input is the noise matrix $z$, which has the following definition:
\begin{equation}
	\begin{aligned}
	z = \{ z_{ij} \leftarrow  \mathcal{N}(0,1),\: \forall & i\in [1,n],\\
							& j\in [1,l],\:i,j\in \mathbb{N}   \}
	\end{aligned}
	\label{eq:gan_noise_sampling}
\end{equation}
where $n$ is the number of samples to be generated (in this case $b/2$) and $l$ is the latent dimension of the generator. Basically, a noise matrix is sampled from the normal distribution.

The second input are the one-hot vectors of the classes $\iota$ of the samples to be generated, which are sampled from a discrete uniform distribution:
\begin{equation}
	\iota = \left\{ \iota_i \leftarrow \mathcal{U}\{1,n_c\},\  \forall i \in [1,n] \subset \mathbb{N} \right\}
	\label{eq:gan_index_sampling}
\end{equation}
where $n_c$ is the total number of classes and $n$ is the number of samples. The one-hot encoded input is $\phi (\iota)$. For the stochastic targets, the target vector is a vector where the element $\bm{t}_{k=\iota}$ has a certain value $p'$ between 0 and 1, while the other elements sum up to 1:
\begin{equation}
	\bm{t}_k =
	\begin{dcases}
		p', & k=\iota \\
		\frac{1-p'}{n_c - 1}, & k\neq\iota
	\end{dcases}
	\label{eq:gan_noisy_target}
\end{equation}

Until now, we have defined all the data required to train the discriminator. There are two analogue tuples of data: $(x_r,t_r,s_r)$ and $(x_g,t_g,s_g)$. The samples $x_r$ and $x_g$ are fed into the discriminator $D$:
\begin{equation}
	(v,y) = D(x)
\end{equation}
There are now two losses to be calculated, as seen at the end of figure \ref{fig:gan_custom_acgan_training}: the validity and classification loss. The validity is a binary classification problem, thus, the loss function chosen is the binary cross-entropy:
\begin{equation}
	L_1 = -{(s\log(v) + (1 - s)\log(1 - v))}
	\label{eq:gan_validity_loss}
\end{equation}
where $L_1$ is the validity loss on a sample, $s$ is the value of sample's source (0 or 1) and $v$ is the predicted probability of the sample belonging to the correct class, which is the first output of $D$.

The actual classification of a sample is a multi-class problem, so the multi-class cross-entropy is used:
\begin{equation}
	L_2 = -\sum_{c=1}^{n_c} t_{i,c}\log(y_{i,c})
	\label{eq:gan_class_loss}
\end{equation}
where $L_2$ is the classification loss of a sample $i$, $t_{i,c}$ is the value of element $c$ of the sample's target $t$, and $y_{i,c}$ is the output of $D$ for the same sample.

The final loss is a weighted average of $L_1$ and $L_2$. Finally, the discriminator's weights are updated.

\subsubsection*{Second stage: generator}

While the discriminator is trained with both real and generated samples, the generator is trained without real samples. The process is the same as shown in figure \ref{fig:gan_custom_acgan_training}, but the data set is not used.

The noise $z$ and indexes $\iota$ are sampled according to (\ref{eq:gan_noise_sampling}) and (\ref{eq:gan_index_sampling}). Analogously to discriminator training, a batch of $b$ samples $x_g$ is obtained from the generator so that $x_g = G(z_g,\phi(\iota_g))$. We then calculate the validity (\ref{eq:gan_validity_loss}) and classification (\ref{eq:gan_class_loss}) losses with the discriminator. Hence, the loss function of the discriminator is also used with the generator. However, the discriminator weights are frozen during this stage, so only the generator's weights are updated to minimize these losses.

\subsection{Classification Decision}

A trained discriminator can be used to classify new samples. However, the output class provided by the discriminator is rarely taken as the final classification. There is often a decision method that provides the final classification, possibly context-based information, such as data from other sensors.

Most of the neural network classifier models have a final layer which is a softmax transfer function. This function provides a probability distribution over the possible classes. This is the second output of $D$ shown in (\ref{eq:gan_d}). The probability of a given sample $x$ belonging to class $i$ is given by:

\begin{equation}
	\bm{y}_i=p\left(  i \mid \bm{x} \right),\quad \text{for}\ i=1,2,...,n_c
\end{equation}
where $n_c$ is the number of possible classes.

Given the probability distribution $\bm{y}$, it makes sense to set a threshold $\tau$ on $\bm{y}_i$ so that if its value drops below a pre-defined value, the output class is disregarded as \textit{others}:
\begin{equation}
	\text{output class} =
	\begin{dcases}
	\text{class}\ i, & \max\ \bm{y} \geq \tau \\
	\text{others}, & \max\ \bm{y} < \tau
	\end{dcases}
	\label{eq:gan_decision_threshold}
\end{equation}
This definition introduces the problem of determining an adequate value for the threshold.

% The definition of a detection threshold $\tau>0$ means that a number of predictions (where $p(y|x) < \tau$) will be classified as novelty, which will cause the gesture classification accuracy (GCA) to drop. Therefore, threshold optimization requires a target accuracy lower than that found when $\tau=0$, which is the reason for the choice of GCA optimization targets ($p=0.90$ and $p=0.85$).

The use of this method results in more false negatives than using no threshold (or $\tau=0$). As a consequence, increasing the threshold yields lower recall of the classes $i$. Therefore, it is possible to define a threshold value such that the recall does not decrease further than, e.g., $5$ or $10\%$. In most applications, false positives are worse than false negatives, the exception being in critical conditions such as a request for an emergency stop. These cases are rare and should be specially handled to increase safety.

\section{Tests and Results}
% What tests are we doing? 
% A. we want to test the GAN implementation + discriminator
% A1: training/validation stats (loss/accuracy)
% A2: sample similarity analysis between data set and generated, and !noise!
% A3: sample representations (drawings)
% B: we want to evaluate the accuracies of the discriminator networks
% B1: accuracy, precision and recall in the 4 tests, comparison
% B2: binary and class accuracies without score threshold, at 95% and 90% class accuracy, plots
%
% Define new class "others"
% Define data sets and data splits
%
% Repeat for MYO and CG

This section describes the test methodologies followed in this chapter. We are interested in evaluating the performance of both the generator and the discriminator. The generator should find patterns in the data set samples and create new samples based on those patterns.

\subsection{Result Validation}

% Data splits' purpose
All tests use the same data split: 60\% for the training set, 20\% for validation and 20\% for testing. The split was fixed at the beginning of analysis, so that it is not optimized for proposed methodology and no data leakage occurs. We consider this hold-out split to be better than a k-fold cross-validation split in regard to deployment-oriented analysis. On one hand, a k-fold method requires the classifier to be trained $k$ times, thus having a training time roughly $k$ times larger than a hold-out split. On the other hand, \acp{GAN} are remarkably difficult to train and it is highly unlikely that the same hyperparameter set will allow the \ac{GAN} training process to converge in all folds.

The training set is used to train the classifier and does not include any sample of non-gestures. The purpose of the validation set is to optimize the hyperparameters of the classification model, i.e., neural network structure, added noise, normalization, among others. Additionally, the model fitting process is controlled by the model loss that is calculated online on the validation set, in order to prevent over-fitting on the training data. This set does not include any non-gesture sample, in order to prevent leakage of these data into the fitting process. Finally, the test set is used to test the generalization capability of the model and is only used when the training and validation sets provide desirable metrics. Therefore, the model is not optimized for the test set and the metrics calculated on it should provide a good measurement of the model performance in other conditions.

\subsection{Data Sets} \label{subsec:gan_datasets}
%\subsection{UC2018 DualMyo Data Set}

We tested the presented methodology on two data sets: the UC2017 Static and Dynamic Hand Gestures data set \cite{UC2017} and UC2018 DualMyo data set \cite{UC2018Myo}. In the experiments, we denote the index 0 to the network abstractions for the first data set ($GAN0$, $D0$, $G0$ {\color{black} for the GAN, D and G of the first data set, respectively)} and 1 for the second ($GAN1$, $D1$, $G1$).

%% UC2017 SGs
% Data set definition (summary)
The UC2017 data set contains static and dynamic gesture samples captured with a data glove and a magnetic tracker. The library contains 24 static gesture classes with samples obtained from 8 subjects with a total of 100 repetitions for each of the 24 classes (2400 samples in total). The classifier is trained on 19 classes and the remaining 5 were set aside to be used as the \textit{others} class (novel patterns). 

% Feature definition
There are no particular features extracted from these data. The networks are trained with raw data, which includes the hand's joint angles provided by the data glove and the hand's pitch, sensed by the tracker. Thus, the features chosen are simply a subset of channels of the available data. Finally, the features are standardized by $x'_{i}=\left(x_{i}-\bar{x}_{i}\right)/s_{i}$, where $x'_{i}$ is the standardized value of feature $i$, $x_{i}$ is the value of the feature,  $\bar{x}_{i}$ and $s_{i}$ are the mean and standard deviation of the feature in the training set. The validation and test sets are standardized by the same parameters. 

%% UC2018 DualMyo
% Data set definition (summary)
The UC2018 DualMyo data set comprises 8 classes of patterns with 110 repetitions each. Class 7 was set aside to become the class \textit{others}, or novel patterns. This means that the classifier is trained on classes 0 through 6, and tested on all 8 classes. This class was selected because it is not trivially separated from the others in an unsupervised manner.

% Feature definition
Data samples are matrices $\bm{X}\in\mathbb{M}^{t\times d}$, where $t$ {\color{black} is the sample length} (200 frames) and the second dimension $d$  consists of the 16 \ac{EMG} channels. The feature extraction function chosen $\mathcal{F}$ is the standard deviation of the sample along time, i.e., one standard deviation per channel. Therefore, the feature vector extracted from each sample $\bm{f}^{(i)}$ is a vector of length equal to the number of channels. This feature is often proportional to muscle contraction strength, thus providing a muscle activation map around the forearm.

%The 60/20/20 split results in 462 samples being used for training, 154 for hyper-parameter validation and 155 for testing. Additionally, the 110 samples of class 7 constitute a second test split to evaluate the performance of the nets on the class \textit{others}.

% GAN structure (discriminator + generator)
\subsection{GAN Structure}

\begin{figure*}
	\centering
	\def\svgwidth{\linewidth}
	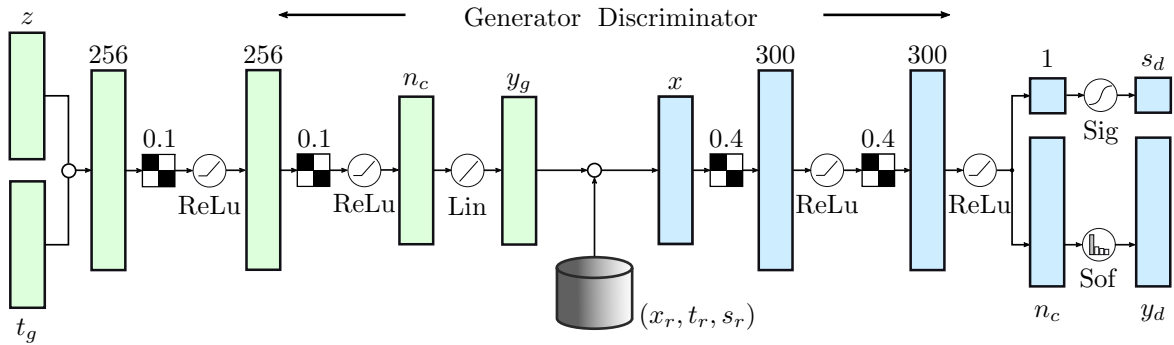
	\caption{Structure of the GAN used on the UC2018 DualMyo data set for the evaluation of the proposed methodology.}
	\label{fig:gan_myo_gan_structure}
\end{figure*}

In the presented framework, the discriminator and generator networks may take many shapes. Furthermore, there is no method to initialize a network structure for a given problem. Generally, it depends on the size of the data set, type and quality of the data and number of features, among other factors. The initial structure is found by random grid search on a varying number of layers and respective nodes in large steps. The network's performance is evaluated on the validation set. When a good structure is found, it is then optimized by manually fine-tuning the number of nodes, transfer functions and applying generalization aids. In this study, we fixed the structure of the GAN's networks for all experiments.

The structure of the generator is shown on the left side of figure \ref{fig:gan_myo_gan_structure}. The depth of the network was kept at two fully-connected (dense) hidden layers with 256 nodes each. There are two inputs, the first being a noise vector $z$ sampled from a random normal distribution $z\sim \mathcal{N}(\mu=0,\sigma=1)$, with a latent size equal to the number of features $n_f$. The second input is the stochastic target vector. These are fully-connected to the first dense layer. As shown in figure \ref{fig:gan_myo_gan_structure}, Gaussian noise $\mathcal{N}(0,0.1)$ is added to each of the dense layers, followed by rectified linear units (ReLU). The ReLU activations are  normalized in the training batch, i.e., centred and scaled. Finally, the output is reduced to a vector of length $n_f$ by a dense layer of the same size, after which a linear transfer function provides the network output. The length $n_f$ corresponds to the number of features of a data set sample, so the output is equivalent to the features of a real sample.

The introduction of Gaussian noise layers is typically recommended in generator networks and helps increase the variance of generated samples. A failure mode of \acp{GAN} is the mode collapse, in which the generator learns and outputs a unique sample. Adding noise during the training process, whether through noise layers or noisy labels, helps prevent this issue.

%\begin{figure*}
%	\centering
%	\def\svgwidth{\linewidth}
%	\input{image_gan_myo_gen_structure.pdf_tex}
%	\caption{Structure of the generator used on the UC2018 DualMyo data set for the evaluation of the GAN methodology.}
%	\label{fig:gan_myo_gen_structure}
%\end{figure*}

The discriminator is shown on the right side of figure \ref{fig:gan_myo_gan_structure}. It has a single input $x$, which can be either generated ($y_g$) or real ($x_r$) samples, therefore having length $n_f$. A Gaussian noise layer $\mathcal{N}(0,0.4)$ is also added. Following that, there is a dense layer with 300 nodes and a ReLU activation function. This layer is repeated one more time with the same parameters. To aid generalization, there is a dropout layer before the output, where $30\%$ of randomly chosen connections are dropped in each training iteration. Afterwards, the network splits into two branches corresponding to its two outputs. The first output, the validity $s_d$, has a single fully-connected node, whose activation is transformed by the sigmoid function in order to return a value between 0 and 1. The second output is also fully-connected with $n_c$ nodes, which corresponds to the number of classes of the problem. In this last case, a softmax activation function is applied in order to output a distribution of probabilities over the number of classes.

%\begin{figure*}
%	\centering
%	\def\svgwidth{\linewidth}
%	\input{image_gan_myo_discr_structure.pdf_tex}
%	\caption{Structure of the discriminator used on the UC2018 DualMyo data set for the evaluation of the \ac{GAN} methodology.}
%	\label{fig:gan_myo_discr_structure}
%\end{figure*}

\subsection{Training Parameters} \label{subsec:gan_training_param}

In section \ref{subsec:gan_training}, we mentioned that the discriminator $D$ and generator $G$ are trained separately, one after the other, since they are different networks. Therefore, they may have distinct training parameters, and the success of a \ac{GAN} framework is strongly dependant of these parameters. The learning process must be balanced so that one does not learn much faster than the other, causing a mode collapse failure. This occurs when the generator is trained to a state where it always outputs the same value, independently of the input, thus bringing the learning process to a halt.

The networks are balanced through the tuning of the learning rates of $D$ and $G$, and the relative weights of the two losses of the generator. As a reminder, the generator is trained with the losses calculated on the discriminator. These are the validity (\ref{eq:gan_validity_loss}) and classification losses (\ref{eq:gan_class_loss}). If the classification losses decrease faster than the validity, the generator will focus on generating $n_c$ different classes that are more easily separable, not necessarily resembling the original data. Therefore, we increase the validity loss weight, so that $G$ learns to generate samples more akin to real data, rather than more separable data. 

Other relevant parameters include the training momentum, number of epochs, the latent dimension of $G$, batch size and label noise strength. Additionally, there are the learning rates for $D$ and $G$, and the loss weights of $D$. We can select recommended (non-optimal) values for these parameters, but since there is no established optimization process, they were optimized by trial and error as a function of the network losses and sample visualization.

Despite the GAN structure being the same for both the UC2017 SG and UC2018 DualMyo data sets, the training parameters were optimized in different directions. The networks $GAN0$ and $GAN1$ were trained for 600 and 300 epochs, respectively, while the batch size was 32 samples in both cases. The $G0$ and $G1$ latent sizes was set to 23 and 8. The stochastic target labels were sampled from $\mathcal{U}(0.9,1.0)$. In early trials, it was set to $\mathcal{U}(0.8,1.0)$, but the lower bound of $0.9$ achieved better results in combination with the remaining parameters.

The \ac{SGD} optimization algorithm followed the Adam update rule \cite{Kingma2014a}. The learning rates were set firstly to the typical value of $0.0002$ for both networks and data sets, but the rate for $G$ ended up being incrementally increased to $0.001$, to speed up the generator training. The discriminators $D0$ and $D1$ were trained with learning rates of $0.001$ and $0.0002$, respectively, and the momentum was kept at 0.5 in all cases. For both $GAN0$ and $GAN1$, the weight decay was set to $10^{-7}$ and $10^{-6}$ to $D$ and $G$, respectively. Finally, the $G$ validity loss weights were set to $1.1$ and $1.3$ for $G0$ and $G1$, respectively, while the classification loss weights were $1.0$ for $G0$ and $0.8$ for $G1$. The training time for this setup is about 289 seconds in a Tensorflow-based Keras implementation for $GAN0$ (63 seconds for $GAN1$). The nets are trained on a Nvidia GTX970M GPU with 6GB of memory.

\begin{figure}
	\centering
	\includegraphics[width=\columnwidth]{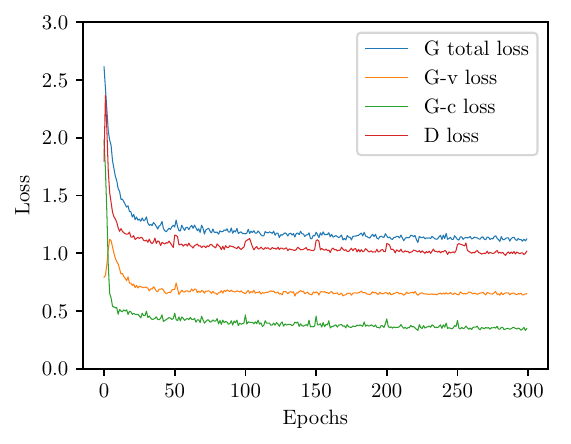}
	\caption{Plot of the training losses of discriminator $D1$ and generator $G1$ validity loss ($G$-v) and classification ($G$-c) loss components for each training epoch. All losses are monotonically decreasing and the $G$-v loss has plateaued by epoch 300.}
	\label{fig:gan_myo_losses}
\end{figure}

An example of the GAN1 losses is shown on figure \ref{fig:gan_myo_losses}. The discriminator and generator losses are very close, which is expected since the loss functions are the same, despite the $G1$ loss being calculated from generated samples, while the $D1$ loss is obtained from equal amounts of generated and real samples. Both losses are still decreasing by epoch 300. However, looking at the generator loss components individually, we see that the classification loss is decreasing, but the validity loss has plateaued. This means that the network is improving the separation of the generated samples but their similarity to the real samples is not improving. The slight peaks found every 50 epochs are likely to be numerical errors caused by sampling $G1$ at the checkpoints that occur at the same frequency.

\subsection{Generator Performance}

Firstly, we tested the quality of the samples created by the \ac{GAN}'s generator. The quality is determined by the similarity between generated and real samples. There are plenty of similarity measures that can be used, but since we have small feature vectors, we opted by simply using the L2 distance between samples. The concept of distance is opposite to similarity, thus the distance must be minimized to improve similarity.

Formally, we are interested in knowing the distance between two sets of samples, $\bm{X}$ and $\bm{Y}$. These sets of data are matrices of shape $(n_{samples}\times n_{f})$, where $n_{samples}$ may or may not be the same. The L2 distance between the $i$-th sample of $\bm{Y}$ and $\bm{X}$ is thus given by:
\begin{equation}
\bm{l}_i = \frac{1}{N}
\sum_{j=1}^{N} \sqrt{ 
	\sum_{k=1}^{n_f} \left( \bm{X}_{jk} - \bm{Y}_{ik} \right)^2 }
\label{eq:gan_g_distance}
\end{equation}
where $N$ is the number of samples in $\bm{X}$ and $n_f$ the number of features. In short, the distance between a sample $\bm{Y}_i$ and the set $\bm{X}$ is the mean L2 distance between $\bm{Y}_i$ and all of the samples of $\bm{X}$. \\

The following tests are presented:
\begin{enumerate}
	\item \textbf{Data set baseline distance:}\\
	Mean distance between data set samples of the same class.\\
	Standard deviation of the distance between data set samples of the same class.
	\item \textbf{Generated data distance:}\\
	Mean distance between real and generated samples of the same class. \\
	Standard deviation of the distance between real and generated samples of the same class.
	\item \textbf{Gaussian noise distance:}\\
	Mean distance between real and random noise generated from Gaussian distributions.\\
	Intra-class standard deviation of the distance between real and noise samples.
\end{enumerate}
Two baseline distances are established: a data set and a random noise baseline. The data set baseline establishes the ideal distance, i.e., the dispersion of the real data set. The generated data distance measures how far the generated data are from the real data, which should tend to the ideal metric when these mimic perfectly the real data. The Gaussian noise distance is the worst-case scenario that would occur if the generator were to diverge from the real data. In all cases, the distance metric is computed for sets of samples within the same class. The randomly generated data are sampled from Gaussian distributions with mean and standard deviations calculated from the real data for each class individually.

% It is also possible to obtain a more intuitive perception of similarity between samples by plotting them.

\begin{figure*}
	\centering
	\includegraphics[width=\linewidth]{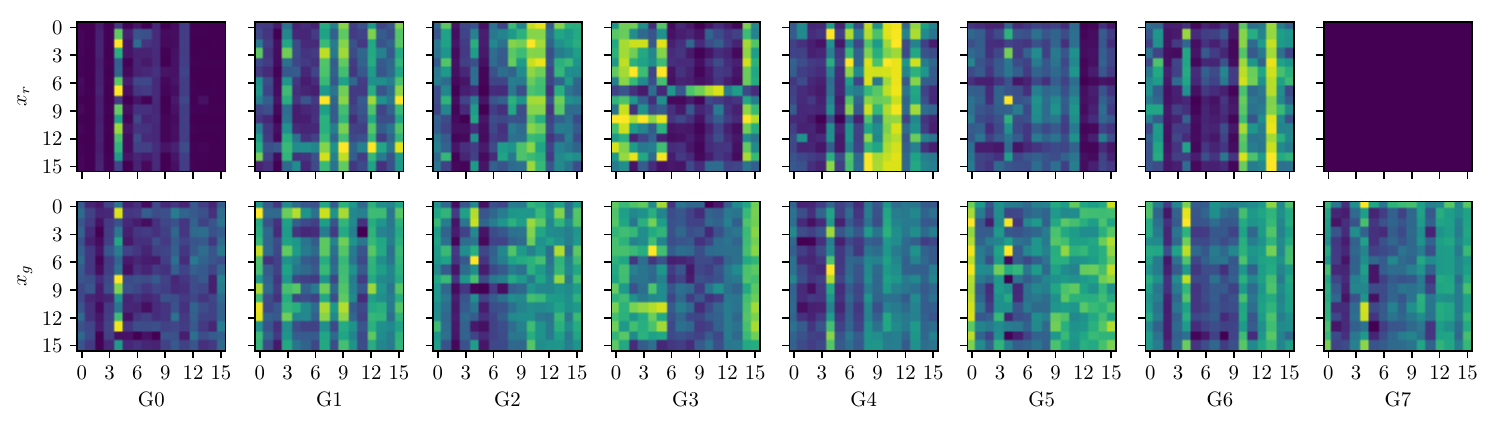}
	\caption{Comparison between real ($x_r$) and generated samples ($x_g$) of each class of the UC2018 DualMyo. G7 is a class created by the \ac{GAN}, so it does not have a real class equivalent.}
	\label{fig:gan_myo_samples}
\end{figure*}

%%% OLD SUB-SECTION FOR DUALMYO

Examples of generated and real samples for each gesture class of the DualMyo data set are shown in figure \ref{fig:gan_myo_samples}, where the lowest signal is represented in dark black and the highest in yellow. A first look shows that generally, the intensity of low state signals is higher in generated samples than in real ones. Nevertheless, this behaviour is not an issue since there is still a visible gap between low and high signal states. The last class, G7 or \textit{others}, is a class created by the generator that was not trained, therefore there is no equivalent in real samples.  

While visualizing the samples provides a subjective evaluation of the quality of the generated samples, the first generator test is comprised by a similarity measurement between data set samples, generated samples and samples drawn from Gaussian distributions. The results are shown in table \ref{tab:gan_myo_gen0_similarity} for the UC2017 SG data set and table \ref{tab:gan_myo_gen1_similarity} for the DualMyo data set. The mean distance values displayed in table \ref{tab:gan_myo_gen0_similarity} show that, by class, most of the GAN-generated samples are significantly closer to the baseline than random samples. The exceptions are classes 0, 2, 3 and 17. However, visualizations of the samples show that the generated samples are similar to the real samples. The dispersion of the GAN-generated samples is closer to the baseline than random, in all cases. The low dispersion of random samples is uncharacteristic of real samples, and might explain why their mean distance is lower than GAN samples in some cases.

In respect to the DualMyo data set, table \ref{tab:gan_myo_gen1_similarity} shows that \ac{GAN}1 is significantly better than the random distribution in mean distance, for all classes. However, the standard deviation of the distance, which measures the dispersion within samples of the same class, is generally significantly lower than that of the data set. While the dispersion is comparatively low, it is significant, as seen in figure \ref{fig:gan_myo_samples}. It also indicates that the training of the \ac{GAN} has not collapsed into generating a single type of sample.

Owing to the similarity between generated and trained samples, and the reasonable intra-class dispersion of the generated samples, we can conclude that the generators are successfully trained.

%\begin{table*}[]
%	\centering
%	\caption{Similarity performance indicators between samples of the data set, Gaussian noise samples and \ac{GAN}-generated samples. The mean and standard deviation of the distances between samples is shown for each class of gestures. The difference $\Delta$ between values and the baseline is shown as a percentage. }
%	\label{tab:gan_myo_generator_similarity}
%\begin{tabular}{@{}ccc *4{cS[table-format=3.2]}@{}}
%	\toprule
%	& \multicolumn{2}{c}{Baseline} & \multicolumn{4}{c}{GAN} & \multicolumn{4}{c}{Noise} \\ \cmidrule(rl){2-3}\cmidrule(rl){4-7}\cmidrule(rl){8-11}
%	Class & Mean & Std & Mean & {$\Delta (\%)$}  & Std & {$\Delta (\%)$} & Mean & {$\Delta (\%)$}  & Std & {$\Delta (\%)$}  \\ \midrule
%	0 & 1.43 & 0.63 & 1.44 & +1.0 & 0.33 & -47.1 & 4.74 & +232.7 & 0.20 & -67.7 \\
%	1 & 3.20 & 1.40 & 3.55 & +11.0 & 0.86 & -38.9 & 5.06 & +58.2 & 1.58 & +12.8 \\
%	2 & 2.28 & 0.69 & 2.10 & -7.8 & 0.57 & -17.3 & 2.47 & +8.4 & 0.58 & -15.6 \\
%	3 & 4.03 & 1.66 & 4.09 & +1.5 & 0.93 & -44.1 & 6.36 & +57.8 & 1.89 & +14.1 \\
%	4 & 2.32 & 0.71 & 2.18 & -6.0 & 0.63 & -11.3 & 4.02 & +73.1 & 0.70 & -1.7 \\
%	5 & 1.95 & 0.77 & 2.00 & +2.2 & 0.85 & +9.7 & 2.55 & +30.6 & 0.46 & -40.2 \\
%	6 & 2.44 & 0.83 & 2.38 & -2.5 & 0.81 & -2.8 & 3.24 & +32.8 & 0.47 & -44.0 \\ \bottomrule
%\end{tabular}
%\end{table*}

% SG/DG
\begin{table}[]
	\centering
	\caption{Similarity performance indicators between samples of the UC2017 SG data set, Gaussian noise samples and \ac{GAN}-generated samples.}
	\label{tab:gan_myo_gen0_similarity}
	\begin{tabular}{@{}*7{c}@{}}
		\toprule
		& \multicolumn{2}{c}{Baseline} & \multicolumn{2}{c}{GAN} & \multicolumn{2}{c}{Random} \\ \cmidrule(rl){2-3}\cmidrule(rl){4-5}\cmidrule(rl){6-7}
		Class & Mean & Std & Mean & Std & Mean & Std \\ \midrule
		0 & 4.32 & 1.83 & 5.06 & 2.41 & 4.49 & 1.60 \\
		1 & 4.19 & 1.45 & 3.96 & 1.38 & 4.85 & 0.77 \\
		2 & 4.33 & 1.54 & 4.19 & 1.41 & 4.45 & 0.97 \\
		3 & 4.56 & 1.84 & 3.95 & 1.20 & 4.22 & 1.52 \\
		4 & 2.84 & 1.01 & 2.58 & 0.96 & 3.70 & 0.76 \\
		5 & 4.57 & 1.85 & 4.29 & 1.66 & 6.35 & 1.07 \\
		6 & 4.82 & 1.72 & 4.32 & 1.49 & 6.27 & 1.36 \\\bottomrule
	\end{tabular}
\end{table}

% DUALMYO
\begin{table}[]
	\centering
	\caption{Similarity performance indicators between samples of the UC2018 DualMyo data set, Gaussian noise samples and \ac{GAN}-generated samples.}
	\label{tab:gan_myo_gen1_similarity}
	\begin{tabular}{@{}*7{c}@{}}
		\toprule
		& \multicolumn{2}{c}{Baseline} & \multicolumn{2}{c}{GAN} & \multicolumn{2}{c}{Random} \\ \cmidrule(rl){2-3}\cmidrule(rl){4-5}\cmidrule(rl){6-7}
		Class & Mean & Std & Mean & Std & Mean & Std \\ \midrule
		0 & 1.43 & 0.63 & 1.44 & 0.33 & 4.74 & 0.20 \\
		1 & 3.20 & 1.40 & 3.55 & 0.86 & 5.06 & 1.58 \\
		2 & 2.28 & 0.69 & 2.10 & 0.57 & 2.47 & 0.58 \\
		3 & 4.03 & 1.66 & 4.09 & 0.93 & 6.36 & 1.89 \\
		4 & 2.32 & 0.71 & 2.18 & 0.63 & 4.02 & 0.70 \\
		5 & 1.95 & 0.77 & 2.00 & 0.85 & 2.55 & 0.46 \\
		6 & 2.44 & 0.83 & 2.38 & 0.81 & 3.24 & 0.47 \\\bottomrule
	\end{tabular}
\end{table}

\subsection{Discriminator Performance}
% 4 tests: baseline, baseline + noisy labels, gan discriminator, gan discr + new generated samples (say how many)
% in each test: no threshold applied, threshold optimized for p_class=0.95, threshold optimized for p_class=0.90
% conclude that noisy labels allow smaller thresholds and higher accuracy
% must do new confusion tables with precision, recall and accuracy

The objective of the presented methodology is to improve the real-world performance of the discriminator. The performance is strongly tied to the accuracy of the classifier model, i.e., the number of successful classifications over the total number of gesture samples. However, the accuracy does not reflect the rate of non-gestures classified as gestures, which may happen in real-world conditions. The metric of interest for this type of problem is the prediction accuracy.

The analysis of the discriminator's performance is done as a two-step problem. The first step is the binary classification problem of novelty detection, i.e., determining whether a new sample belongs to one of the trained classes or not. {\color{black} The relationship between the true positive rate (correctly detected novel samples) and false positive rate is provided by the receiver operating characteristic (ROC) curve.}  Models are compared by their area under curve (AUC) in the ROC curve. The second problem is multi-class classification, where a prediction is performed to find what the class of a sample is, within the trained classes subset. The chosen performance metric is the classification accuracy (\ref{eq:gan_accuracy_metrics}). The novel detection accuracy (NDA) is defined as the fraction of novel samples that are correctly identified. On the other hand, gesture classification accuracy (GCA) is the fraction of correctly discriminated samples that belong to new classes.

%Define accuracy
\begin{equation} 
	\label{eq:gan_accuracy_metrics}
	\text{Accuracy}=\frac{\text{Number of correct predictions}}{\text{Total number of predictions}}
\end{equation}

% where TP, FP and FN stand for true positive, false positive and false negative, respectively. In the multi-class problem, the TPs are considered to be the cases when a pattern of class $\tau$ is classified as such and FPs occur when $\tau$ is classified as any other class.

%Enumerate tests
The working hypothesis is that the data set augmentation performed with the previously described \acp{GAN} improves the classification of gestures and the unsupervised classification of novel gestures. The baseline performance is established by training a discriminator model (baseline A) with the data set samples. {\color{black} For dynamic gesture classification (UC2018 DualMyo data set), a gated recurrent unit (GRU) \ac{RNN} was used as a second baseline B.} Following that, since we found that presence of stochastic labels on the data set helps training the \ac{GAN}, we test their specific contribution without using the \ac{GAN} framework. Finally, we test the performance of the discriminator trained within the \ac{GAN} framework and retrained with real and generated samples:
\begin{description}
	\item \textbf{Baseline A:} Discriminator trained regularly;
	\item {\color{black} \textbf{Baseline B:} \ac{RNN} discriminator trained regularly;}
	\item \textbf{Test 1A:} Discriminator trained with stochastic target vectors;
	\item {\color{black}  \textbf{Test 1B:} \ac{RNN} discriminator trained with stochastic target vectors;}
	\item \textbf{Test 2:} Trained \ac{GAN} discriminator (online augmentation);
	\item \textbf{Test 3:} Trained \ac{GAN} discriminator retrained with real and generated samples (offline augmentation).
\end{description}

Additionally, these tests are repeated without the threshold defined in (\ref{eq:gan_decision_threshold}) and with the threshold tuned so that the GCA is $95\%$, $90\%$ or $85\%$. The tests were performed on the UC2017 SG and UC2018 DualMyo data sets, described in section \ref{subsec:gan_datasets}. 

The discriminator performance was measured in several settings. To ensure that the differences between them are due to the proposed methodology, the structure of the discriminator, figure \ref{fig:gan_myo_gan_structure}, is fixed for all tests. Additionally, all networks are initialized with the same weights and the data set splits (training, validation and testing) are fixed as well.

The baseline tests consist of the performance measurements calculated with the outputs of a discriminator that was trained as a regular neural network. The training hyperparameters were optimized for this purpose. The Adam optimizer was used with a learning rate of $0.01$ {\color{black}  and $0.005$ for baselines A and B, respectively.} The training process is halted using the early stopping technique, where the loss on the validation data is monitored online. The training process is stopped when the validation loss stops decreasing for 12 consecutive epochs in order to prevent over-fitting on the training data. Since this is a rather small training data set in a simple networks, the fitting processes {\color{black}  take less than a minute}.

In Tests 1A{\color{black}  /1B}, the discriminator performance tests are similar to the baselines, but the training targets are stochastic vectors, as described in (\ref{eq:gan_noisy_target}). The maximum value of the target vector of each sample was sampled before training from a uniform distribution $\mathcal{U}{\sim}(0.8,1.0)$ for A {\color{black}  and $\mathcal{U}{\sim}(0.9,1.0)$ for B}. 

Tests 2 and 3 use the trained \ac{GAN} with the setup described in section \ref{subsec:gan_training_param}. For Test 2, the discriminator is used as trained in the \ac{GAN} framework. For Test 3, the same model is retrained with stochastic labels and the training set is augmented offline with an increase of $50\%$ in the number of samples.

%% ACCURACY RESULTS

\begin{figure}
	\centering
	\includegraphics[width=\columnwidth]{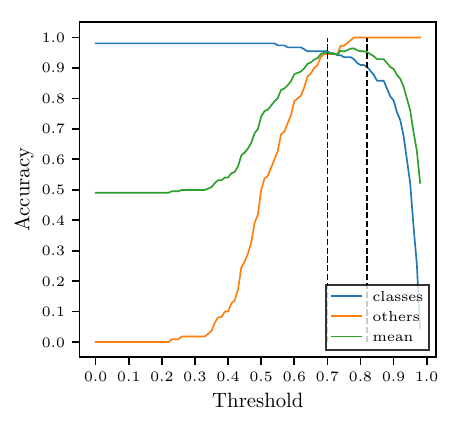}
	\caption{Trade-off between test split GCA and NDA a function of the decision threshold, on the UC2018 DualMyo data set. The two vertical lines correspond to the $p=0.95$ and $p=0.90$ thresholds.}
	\label{fig:gan_myo_d_acc_th}
\end{figure}

\begin{figure}
	\centering
	\includegraphics[width=\columnwidth]{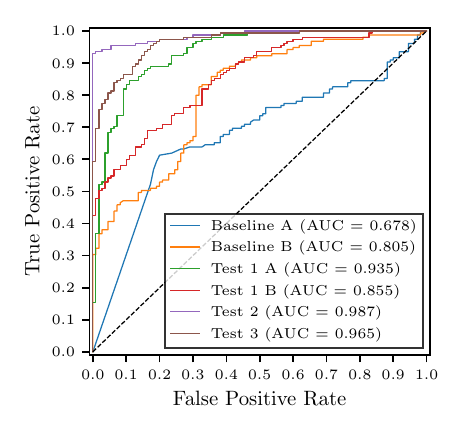}
	\caption{\color{black} Receiver operating characteristic curves of the novelty detection binary classification model for all tests on the UC2018 DualMyo data set.}
	\label{fig:gan_roc}
\end{figure}

All tests were repeated with different values of thresholds $\tau$ for the final classification decision, as defined in (\ref{eq:gan_decision_threshold}). The no-threshold case is equivalent to setting $\tau=0$. An example of the threshold optimization process on the UC2018 DualMyo data set is shown on figure \ref{fig:gan_myo_d_acc_th}. This figure shows that when no decision threshold is set, the GCA is at its maximum while the NDA is at its minimum. As the threshold increases, the GCA (classes) decreases as expected, but the NDA (others) increases faster. Therefore, it is possible to numerically find an optimum balance between GCA and NDA. This balance was roughly set in the two data sets as the maximum GCA attained in the baseline test minus 5\% and 10\%. For example, if the maximum GCA is 100\%, the threshold is tuned so that the GCA is at least 95\% or 100\%.

% TABLE GAN0
\begin{table*}[]
	\centering
	\caption{Accuracy of $D0$'s predictions on the test split. The \textit{Class} and \textit{Others} columns correspond to the trained and \textit{others} classes, respectively.}
	\label{tab:gan_myo_accuracy_sg}
	\begin{tabular}{@{}c *3{S[table-format=3.1]} *3{S[table-format=3.1]}S[table-format=1.2] *3{S[table-format=3.1]}S[table-format=1.2]@{}}
		\toprule
		%& & & & \multicolumn{4}{c}{$p=0.95$} & \multicolumn{4}{c}{$p=0.90$} \\\cmidrule(lr){2-5}\cmidrule(lr){6-9}
		& \multicolumn{3}{c}{$\tau=0$} & \multicolumn{3}{c}{{Accuracy (\%), $p=0.90$}} &  & \multicolumn{3}{c}{{Accuracy (\%), $p=0.85$}} &  \\\cmidrule(lr){2-4}\cmidrule(lr){5-7}\cmidrule(lr){9-11}
		& {Class} & {Others} & {Mean} & {Class*} & {Others} & {Mean} & {$\tau_{0.90}$} & {Class*} & {Others} & {Mean} & {$\tau_{0.85}$} \\\midrule
		Baseline A & 94.7 & 0.0 & 47.4 & 90.0 & 31.2 & 60.6 & 0.92 & 86.3 & 37.6 & 62.0 & 0.96 \\
		Test 1A & 95.0 & 0.0 & 47.5 & 90.0 & 83.0 & 86.5 & 0.67 & 85.0 & 91.0 & 88.0 & 0.81 \\
		Test 2 & 95.8 & 0.0 & 47.9 & 90.3 & 90.2 & 90.2 & 0.93 & 85.0 & 92.0 & 88.5 & 0.95 \\
		Test 3 & 95.3 & 0.0 & 47.6 & 90.0 & 83.2 & 86.6 & 0.66 & 85.5 & 90.0 & 87.8 & 0.77 \\
		\bottomrule
		\multicolumn{12}{l}{*Closest accuracy value to $p$ after threshold optimization.}
	\end{tabular}
\end{table*}

% TABLE GAN1
\begin{table*}[]
	\centering
	\caption{Accuracy of $D1$'s predictions on the test split. The \textit{Class} and \textit{Others} columns correspond to the trained and \textit{others} classes, respectively.}
	\label{tab:gan_myo_accuracy_dualmyo}
	\begin{tabular}{@{}c *3{S[table-format=3.1]} *3{S[table-format=3.1]}S[table-format=1.2] *3{S[table-format=3.1]}S[table-format=1.2]@{}}
		\toprule
		%& & & & \multicolumn{4}{c}{$p=0.95$} & \multicolumn{4}{c}{$p=0.90$} \\\cmidrule(lr){2-5}\cmidrule(lr){6-9}
		& \multicolumn{3}{c}{$\tau=0$} & \multicolumn{3}{c}{{Accuracy (\%), $p=0.95$}} &  & \multicolumn{3}{c}{{Accuracy (\%), $p=0.90$}} &  \\\cmidrule(lr){2-4}\cmidrule(lr){5-7}\cmidrule(lr){9-11}
		& {Class} & {Others} & {Mean} & {Class*} & {Others} & {Mean} & {$\tau_{0.95}$} & {Class*} & {Others} & {Mean} & {$\tau_{0.90}$} \\\midrule
		Baseline A & 100.0 & 0.0 & 58.5 & 95.5 &  5.5 & 58.1 & 0.93 & 91.0 & 10.0  & 57.4 & 0.93 \\
		\color{black} Baseline B &  \color{black} 98.7 & \color{black} 0.0 & \color{black} 57.7 & \color{black} 96.1 & \color{black} 30.9 & \color{black} 69.0 & \color{black} 0.98 & \color{black} 96.1 & \color{black} 30.9  & \color{black} 69.0 & \color{black} 0.98 \\
		Test 1A    & 100.0 & 0.0 & 58.5 & 95.5 & 70.0 & 84.9 & 0.68 & 92.3 & 74.5  & 84.9 & 0.73 \\
		\color{black} Test 1B    &  \color{black} 98.1 & \color{black} 0.0 & \color{black} 57.4 & \color{black} 95.5 & \color{black} 40.9 & \color{black} 72.8 & \color{black} 0.79 & \color{black} 91.6 & \color{black} 50.9  & \color{black} 74.7 & \color{black} 0.84 \\
		Test 2     &  98.1 & 0.0 & 57.4 & 95.5 & 94.5 & 95.1 & 0.70 & 90.3 & 100.0 & 94.3 & 0.82 \\
		Test 3     & 100.0 & 0.0 & 58.5 & 95.5 & 72.7 & 86.0 & 0.69 & 90.3 & 88.2  & 89.4 & 0.82 \\ \bottomrule
		\multicolumn{12}{l}{*Closest accuracy value to $p$ after threshold optimization.}
	\end{tabular}
\end{table*}

The results of all tests on the UC2017 SG data set are shown on table \ref{tab:gan_myo_accuracy_sg}. When no decision threshold is set, the baseline GCA and NDA are 94.7\% and 0.0\%, respectively. 

The baseline test results show that the discriminator's NDA increases from 0 to 37.6\% while the GCA decreases by about 9\% ($p=0.85$). The use of stochastic labels on Test 1A show a significant improvement of NDA to 83.0\% ($p=0.90,\ \tau=0.67$) and 91.0\% ($p=0.85,\ \tau=0.81$). The augmented discriminator of the GAN, Test 2, yields a further improvement to 90.2\% ($p=0.90,\ \tau=0.93$) and 92.0\% ($p=0.85,\ \tau=0.95$). However, retraining the discriminator on Test 3 shows a decrease in performance, when compared to the discriminator augmented online with GANs.

The discriminator tests were repeated for the UC2018 DualMyo data set and the results are shown in table \ref{tab:gan_myo_accuracy_dualmyo}. The behaviour on this data set is similar to what was seen on the previous data set. The classification accuracy without a decision threshold ($\tau=0$) is above 100.0\%, so the threshold optimization targets were set to $p=0.95$ and $p=0.90$. There is a significant increase in NDA between the baseline and the remaining tests. The NDA increases from 5.5\% in Baseline A to 94.5\% on Test 2 ($p=0.95$), or 100.0\% when the classification decision threshold is optimized so that $p=0.90$. However, similarly to the results on the previous data set, Test 3 shows a large drop in NDA to 72.7\%. 

The results show that in all data sets, the proposed methodology greatly increases NDA without significantly impacting GCA, as seen in tables \ref{tab:gan_myo_accuracy_sg} and \ref{tab:gan_myo_accuracy_dualmyo}. The baseline results show that the discriminator is not capable of detecting novelty in any circumstance. Additionally, the decision threshold does not change substantially after optimization in the baseline tests (0.92 to 0.96). {\color{black} Similarly, even though baseline B shows a very high classification accuracy on the trained classes, the optimized threshold is very high (0.98). These results indicate} that most of the predictions done by the neural network show high scores despite the sample's provenance. The use of stochastic target vectors in place of one-hot vectors lowers the prediction scores while maintaining the network's discriminability. Test 1A shows that it is possible to increase NDA while maintaining a high GCA with thresholds as low as 0.67 on the UC2017SG data set. The online data augmentation enabled by GANs yields a further improvement in  NDA, as demonstrated for all data sets in Test 2. The offline data set augmentation tests (Test 3) show lower NDA than online augmentation. This is probably explained by the lower variance of the samples generated by a fully trained generative network (offline augmentation), when compared to the samples created by the dynamic generative model in the GAN framework (online augmentation). 

{\color{black} The ROC curves of the UC2018 DualMyo models are shown in figure \ref{fig:gan_roc}. This plot shows how the AUC increases from 0.678 to 0.987, from the baseline, using the proposed methodology.}

\section{Conclusion}

In this study we implemented {\color{black} a novel method to solve} the issue of classification of out-of-vocabulary gestures. {\color{black} These gestures} are difficult to detect with a simple classification {\color{black} score} threshold. The proposed solution has two components: (1) the use of a generative model (\ac{GAN}) to augment the data set online with new generated samples, (2) the use of stochastic target vectors to decrease the average prediction score, thus facilitating threshold tuning. Finally, a threshold is set on the prediction score in order to make a final classification.

% To evaluate the proposed approach, we defined a baseline test where a neural network is used as a classification model without augmentation or stochastic labels. Test 1 added stochastic vectors in place of the one-hot targets used on the baseline test. The second test added the \ac{GAN} approach, with its discriminator being used as the classification model. In the third and final test, the previous discriminator is retrained with a $50\%$ augmented training split where the new samples are generated with the \ac{GAN}'s generator and with stochastic labels. The discriminator shape and optimization starting point are kept static in all tests. All tests used a threshold on the prediction score in order to make a final classification. In a first approach, these metrics were calculated when there is no classification threshold. Two other cases used tuned thresholds so that there is a maximum loss of trained class accuracy of $5$ and $10\%$. 

Tests were performed on two data sets to determine the influence of the proposed changes on the discrimination capability of a neural network. The results show that the use of stochastic target vectors improves significantly the novelty detection accuracy while maintaining a high classification accuracy. Furthermore, the online augmentation of the discriminator's training data set with a GAN yields a further improvement. However, while offline augmentation may offer better classification accuracy, it showed worse performance on novelty detection when compared to both online augmentation and stochastic target vectors. 
%The results on the UC2017 SG data set showed a maximum classification accuracy of 95.8\%. Without threshold tuning, the NDA is 0.0\% in all cases. When the threshold was tuned, the NDA increased from 0.0\% to 90.2\% using the GAN framework coupled with stochastic target vectors, which is significantly better than the baseline (31.2\%). The results on the UC2018 DualMyo data set showed a similar behaviour, where the proposed methodology presented a NDA of 94.5\% and a baseline of 30.9\%.

% The tests were performed on the UC2018 DualMyo data set with 7 trained classes and 1 non-trained class (\textit{others}). The baseline results showed that without a threshold, all of the samples of \textit{others} were confused with another class. Considering a tuned threshold, only $17.3\%$ of \textit{others} were excluded correctly. Without a threshold, all of the samples of class \textit{others} were excluded. Considering a threshold tuned to a loss of $5\%$ ($p=0.95$) and $10\%$ ($p=0.90$), the best rejection rate of \textit{others} was $95.4\%$ and $100.0\%$, respectively, with the \ac{GAN} discriminator (test 2).

A major challenge of the proposed methodology is the successful training of the \ac{GAN}. The generative model performance can still be improved in order to generate more diverse samples, even though the performance achieved in this work resulted in an improvement of NDA. Despite the success, further validation of the methodology should be done on richer data sets. Further work should be done on the generation of time-series.

%%%% BIBLIOGRAPHY

%\printbibliography

\end{document}